\newif\ifuseoptimistic
\useoptimistictrue 

\documentclass[]{fairmeta}

\usepackage{microtype}
\usepackage{graphicx}
\usepackage{subcaption}
\usepackage{mathptmx}
\usepackage{booktabs}
\usepackage{hyperref}
\usepackage{algorithm}
\usepackage{algorithmic}
\usepackage{xcolor}
\usepackage{amsmath}
\usepackage{amssymb}
\usepackage{mathtools}
\usepackage{amsthm}

\title{To 2:4 Sparsity and Beyond: Neuron-level Activation Function to Accelerate LLM Pre-Training}

\author{Meghana Madhyastha}
\author{Daniel Haziza}
\author{Jesse Cai}
\author{Newsha Ardalani}
\author{Zhiqi Bu}
\author{Carole-Jean Wu}
\affiliation{FAIR at Meta}

\abstract{ Trainings of Large Language Models are generally bottlenecked by matrix multiplications. In the Transformer architecture, a large portion of these operations happens in the Feed Forward Network (FFN), and this portion increases for larger models, up to $50\%$ of the total pretraining floating point operations. We show that we can leverage hardware-accelerated sparsity to accelerate all matrix multiplications in the FFN, with 2:4 sparsity for weights and v:n:m (Venom) sparsity for activations. Our recipe relies on sparse training steps to accelerate a large part of the pretraining, associated with regular dense training steps towards the end. Overall, models trained with this approach exhibit the same performance on our quality benchmarks, and can speedup training end-to-end by $1.4$ to $1.7\times$. This approach is applicable to all NVIDIA GPUs starting with the A100 generation, and is orthogonal to common optimization techniques, such as, quantization, and can also be applied to mixture-of-experts model architectures.}

\date{\today}
\correspondence{\email{meghanam@meta.com}, \email{dhaziza@meta.com}, \email{carolejeanwu@meta.com}}

\begin{document}

\maketitle

\section{Introduction}
Training large language models (LLM) is computationally expensive: the largest models are trained with more than $10^{25}$ floating point operations (FLOP), most of which are matrix multiplications. Hence, any improvement to matrix multiplication efficiency on GPUs can have an important impact on training speed. One way to do so is by reducing the precision of the computations [\cite{abecassis2025pretraining}]. This work focuses on an orthogonal approach: skipping computations by setting model parameter values to zero with sparsity. 

Large model weights can be trained sparsely --- some parameters can be set to zero without any drop in model accuracy or quality [\cite{frankle2019lotterytickethypothesisfinding}]. Sparsity also appears naturally for model activations. Only a small fraction of a model's activations come with meaningful values, with larger models having a larger sparsity ratio [\cite{li2023lazyneuronphenomenonemergence}]. When using hard activation functions, such as ReLU, we observe more than $90\%$ of the post-activation values can be zeros, without any particular loss. The sparsity pattern for activations is dynamic because the position and existence of zeros depends on inputs.

Both weight and activation sparsity is fine-grained meaning sparsity occurs at the granularity of individual neurons/weights in a matrix. This is in contrast with coarse grained sparsity as seen in mixture of expert models. Fine grained sparsity is more challenging as it requires specialized hardware support to be accelerated, otherwise the zeros are actually multiplied and the matrix multiplication is not faster. Furthermore, even if some computations are skipped, the data movement across the memory hierarchy can incur significant energy overhead which may negate the computation gains. Thus, in order to see real end-to-end speedup, data must be stored in a packed or compressed format, which can incur additional overhead when converting to that specific format.

Nvidia introduced a semi-structured format called $n:m$ sparsity along with hardware support (sparse tensor cores) to efficiently accelerate matrix multiplication computations [\cite{mishra2021accelerating}]. In the $n:m$ format, for every group of $m$ elements at most $n$ are non-zero. Starting with Ampere, only $2:4$ sparse matrix multiplication is supported for bf16 or fp8 operands, resulting in $50\%$ sparse matrices. Theoretically, $2:4$ sparse matrix multiplication doubles the throughput performance compared to dense matrix multiplication. In practice, the speedups are closer to $1.4-1.5\times$ (Section~\ref{sec:exp_microbenchmarks}). Researchers have investigated different techniques to use $2:4$ sparse matrix multiplication to accelerate LLM training. 

However, accelerating training with $2:4$ sparsity is challenging for multiple reasons. First, the pruning mask to convert a matrix to the $2:4$ format shouldn't incur a significant overhead since this will be computed dynamically at every training iteration. Second, dropping zeros should not affect the training dynamics and the continuity of the loss curve. Ultimately, the tradeoffs between different sparsification strategies (activations, weights) in both efficiency and model accuracy have not been yet been studied systematically. 

\begin{figure}[th]
  \begin{center}
  \centerline{\includegraphics[width=0.5\columnwidth]{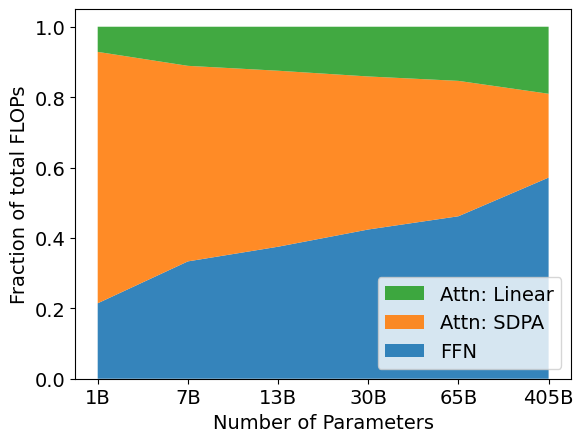}}
    \caption{\centering Fraction of total pretraining FLOPS in each component, as a function of model size. Scaled Dot Product Attention (SDPA) scales linearly with the model dimension, while the linear layers scale quadratically. Hence, for a fixed pretraining sequence length, larger models spend more FLOPS in the Feed Forward Networks (FFNs).}\label{fig:flopfrac}
    
  \end{center}
\end{figure}

In this work, we systematically characterize semi-structured sparsification strategies on  matrix multiplications in the feed forward network (FFN) block in Transformers. We study the effects of different sparsification schemes and derive at a combination/recipe that yields and optimal tradeoff between efficiency and accuracy. We focus on sparsifying the FFN block since it tends to dominate the overall compute (FLOPS), increasingly so at large model sizes as seen in Figure~\ref{fig:flopfrac}. Additionally, the naturally occurring sparsity in the hidden activations is a good fit for activation sparsity. We build on the observation that some activation functions, such as, the SquaredReLU function [\cite{haziza2025acceleratingtransformerinferencetraining}], exhibit a high degree of sparsity (above $90\%$) to explore other formats that can support this sparsity characteristic. Specifically, we explore using the $V:N:M$ (Venom) format since it can accelerate higher ranges of sparsity. However, mapping unstructured sparsity to this format is challenging. We thus introduce a novel activation function that does neuron level MoE style routing to achieve the Venom format. 

Concretely, our contributions are as follows.
\begin{enumerate}
\item We present the first method to sparsify the feed forward networks entirely during training. We do this by applying 2:4 sparsity on weights and Venom sparsity [\cite{Venom}] on activations, such that every matrix multiplication is sparse. 
\item We derive an optimal combination of sparse and dense training steps to match dense pretraining accuracy.
\item We show that this training recipe results in substantial speedups through key microbenchmarks and roofline analysis. 
\end{enumerate}

\section{Background and Motivation}
We start by enumerating the matrix multiplications in the FFN block in Section~\ref{sec:preliminary}. Next, we describe and explain the soft-thresholding technique that we use for weight sparsification in Section~\ref{sec:weight-thresholding}. We then describe the v:n:m (Venom) format in Section~\ref{sec:act}. 

\subsection{Background on Matrix Multiplications in the FFN Block of Transformers}
\label{sec:preliminary}

\begin{figure}[ht]
  \begin{center}
    \centerline{\includegraphics[width=0.4\columnwidth]{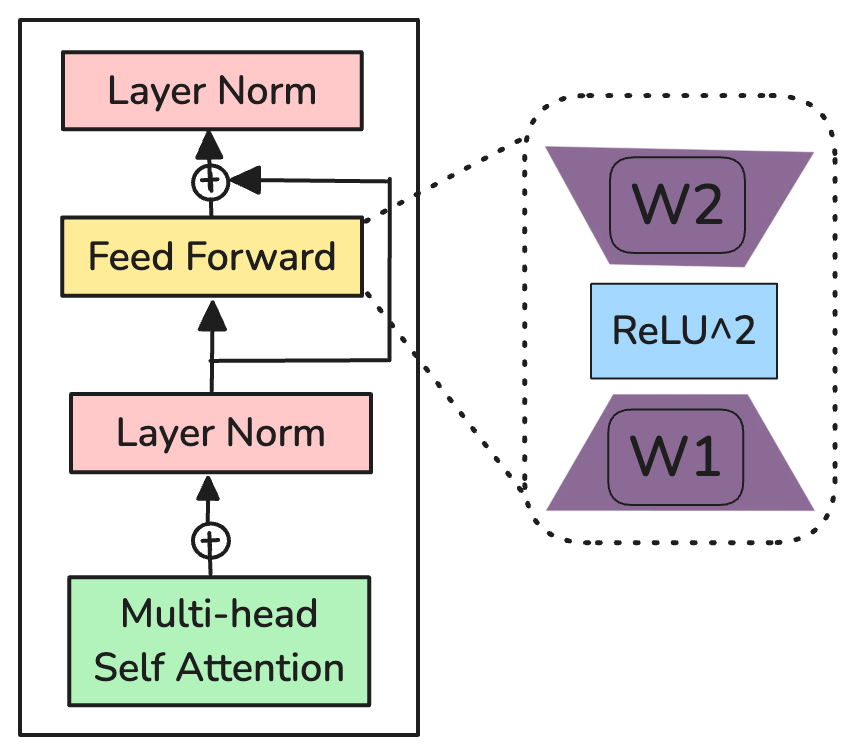}}
    \caption{\centering Transformer Block Diagram.
    }
    \label{block-diag}
  \end{center}
\end{figure}

Figure~\ref{block-diag} shows the block diagram of a Transformer and the FFN block zoomed in. Here are the corresponding matrix multiplications for the forward and backward pass respectively (in the FFNs).\\
\textbf{Forward($x$, $W_1$, $W_2$)}
\vspace{-1.3em}
\begin{align}
y_1 = xW_1 \\
y_2 = \mathrm{ReLU}(y_1)^2 \\
y_3 = y_2 W_2
\end{align}
\textbf{Backward($y_1$, $y_2$, $x$, $W_1$, $W_2$)}
\begin{align}
dy_2 = dy_3 W_2^T \\
dW_2 = y_2^Tdy_3\\
dy_1 = 2dy_2\mathrm{ReLU}(y_1) \\
dx = dy_1W_1^T\\
dW_1 = dy_1^TX
\end{align}
Concretely, Equations (1), (3), (4), (5), (7) and (8) represent matrix multiplications that can be accelerated with 2:4 sparsity. $W_1$ and $W_2$ (corresponding $dW_1$ and $dW_2$) are the weight matrices and their derivatives. $y_2$, $dy_1$ and $dy_2$ are the activation matrices that are naturally more than $90\%$ sparse when using squared ReLU activation function. Training a Transformer using the Squared ReLU has been shown to give comparable results to GeLU based activation functions~\cite{haziza2025acceleratingtransformerinferencetraining}. We thus adopt it since it creates natural sparsity without significantly impacting model quality or accuracy.

\subsection{Weight Sparsification using Soft Thresholding}\label{sec:weight-thresholding}
Naively applying a magnitude based pruning mask results in a discontinuous loss function and thus a degradation in model quality and evaluation performance. To mitigate this issue, we take inspiration from \cite{hu2024sstecontinuouspruningfunction} and apply soft-thresholding that provably removes discontinuity and results in better evaluation performance. Soft-thresholding simply subtracts the 2nd smallest element from each element in groups of 4 (setting the smallest two to zero). Concretely, the soft-thresholding function on a group of four elements: $[a_1, a_2, a_3, a_4]$ is defined as follows.

\[
f(x) = \begin{cases}
    a_i - t, & a_i \in (t, +\infty] \\
    0, & a_i \in (-t, t] \\
    a_i + t, & a_i \in (-\infty, -t]
\end{cases}
\]

Here, $[t_1, t_2, t_3, t_4]$ is a rearrangement of a.T, s.t. $|t_1| \leq |t_2| \leq |t_3| \leq |t_4| $

\subsection{Activation Sparsity using V:N:M Sparsification} \label{sec:act}
\begin{figure}[th]
  \begin{center}
    \centerline{\includegraphics[width=0.5\columnwidth]{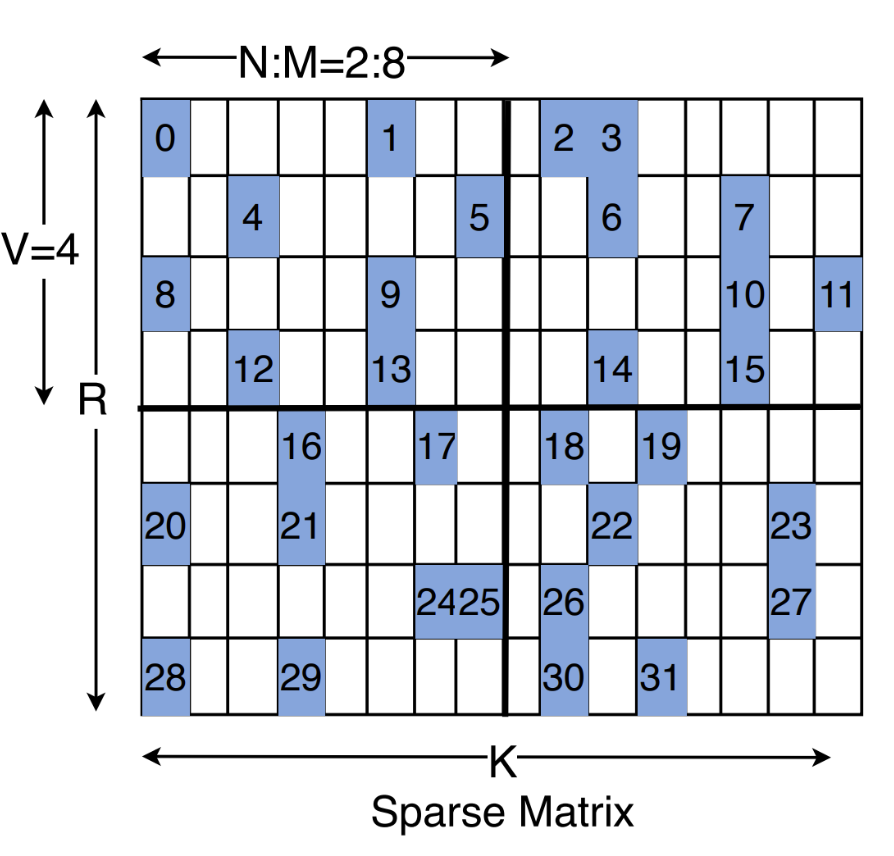}}
    \caption{\centering An illustrating example for a matrix in the Venom format. 
    }
    \label{Venom:basicfig}
  \end{center}
\end{figure}

We observe sparsity levels greater than $90\%$ sparsity in the activations when using the squared ReLU activation function. Hardware accelerated $2:4$ sparsity can only leverage $50\%$ sparsity, whereas the sparsity levels are above $90\%$. This motivated us to rethink the sparsity format. Specifically, we use the $V:N:M$ or Venom format that gives 6-10x speedups for 80-99 percent sparsity levels. Concretely, Venom divides a matrix into sub-matrices of size $[nrows // V, M]$ where $nrows$ is the number of rows in the original matrix, $V$ and $M$ are hyperparameters. In each sub-matrix, only 4 columns are retained (the rest are set to zero). The matrix formed from the 4 columns that remain is then 2:4 sparsified by greedy magnitude pruning. This results in a matrix that is $\frac{N}{N-2}$ sparse.  However, this format isn't purely unstructured since it involves column pruning per sub-matrix followed by 2:4 pruning. Figure~\ref{Venom:basicfig} shows an example of a matrix in the Venom format.

\section{Proposed Design}\label{sec:proposed_design}
Our training recipe accelerates all matrix multiplications in the FFN by ensuring exactly one of their operands is sparse. Activations are sparsified with V:N:M, and the remaining matrix multiplications are accelerated by applying soft-thresholding to 2:4 sparsify the weights. The exact procedure is provided in Algorithm~\ref{alg:fwd} for the forward pass, and in algorithm~\ref{alg:bwd} for the backward pass: the sparse operands are colored in red when sparsified with 2:4 soft-thresholding, and in blue when sparsified with V:N:M.

For some  matrix multiplications, both of the operands could be sparsified, even though hardware accelerated GEMMs are only supported for a single sparse operand. For example, in Equation (3), we could choose to sparsify $W_2$ instead of $y_2$. Similarly, in equation (7), we could choose to sparsify $dy_1$ instead of $W_1^T$. When possible, we prefer activation sparsification. This is because naturally occurring activation sparsity can be better exploited using the Venom format (6x faster GEMMs) compared to the 2:4 sparsification possible with weights (1.5x faster GEMMs).

\begin{algorithm}[ht]
\centering
    \caption{FFN Forward Pass} \label{alg:fwd}
    \begin{algorithmic}[1]
    \STATE $X: [B, d_{model}]$\\
    \STATE $W_1: [d_{model}, d_{FFN}]$\\
    \STATE $W_2: [d_{FFN}, d_{model}]$\\
    \STATE $y_1 = X @ \textcolor{orange}{\operatorname{sparsify24}(W_1)})$ //2:4 matmul\\
    \STATE $y_2 = \mathrm{ReLU}(y_1)^2 $\\
    \STATE $y_2 = \operatorname{MoEToVenom}(y_2)$
    \STATE $y_3 = \textcolor{blue}{y_2} @ {W_2}$ //Venom matmul
    \end{algorithmic}
\end{algorithm}

\begin{algorithm}[ht]
\centering
    \caption{FFN Backward Pass} \label{alg:bwd}
    \begin{algorithmic}[1]
    \STATE $y_2$, $y_1$ -> saved from forward pass
    \STATE $dy_2 = dy_3 @ \textcolor{orange}{\operatorname{sparsify24}({W_2.T})}$ //2:4 matmul\\
    \STATE $dW_2 = \textcolor{blue}{y2.T}@ dy_3$ //Venom matmul\\
    \STATE $dy_1 = 2 \cdot dy_2 \cdot \mathrm{ReLU}(y_1))$
     \STATE $dX =  \textcolor{blue}{dy_1} @ W_1.T$ //Venom matmul\\
      \STATE $dW_1 =  \textcolor{blue}{dy_1.T}@X$ //Venom matmul\\
    \end{algorithmic}
\end{algorithm}

\begin{figure}[th]
    \centering
    \begin{subfigure}[b]{0.48\textwidth}
        \centering
        \includegraphics[width=\textwidth]{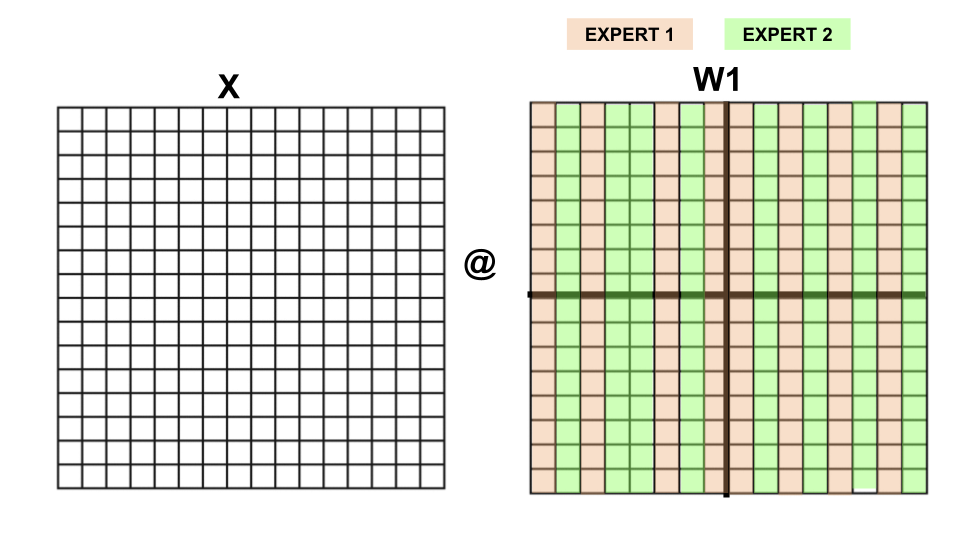}
        \caption{}
        \label{fig:moe1}
    \end{subfigure}
    \begin{subfigure}[b]{0.48\textwidth}
        \centering
        \includegraphics[width=\textwidth]{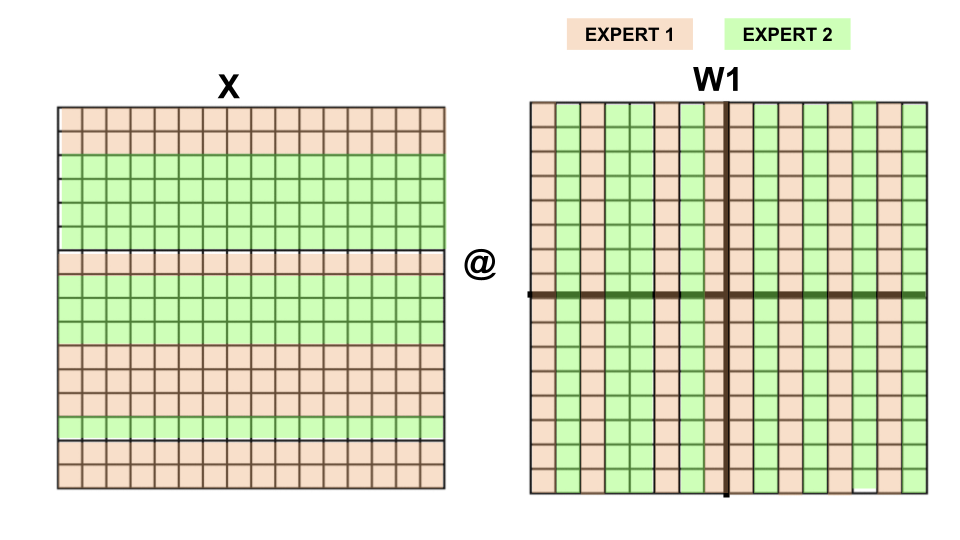}
        \caption{}
        \label{fig:moe2}
    \end{subfigure}
    \begin{subfigure}[b]{0.48\textwidth}
        \centering
        \includegraphics[width=\textwidth]{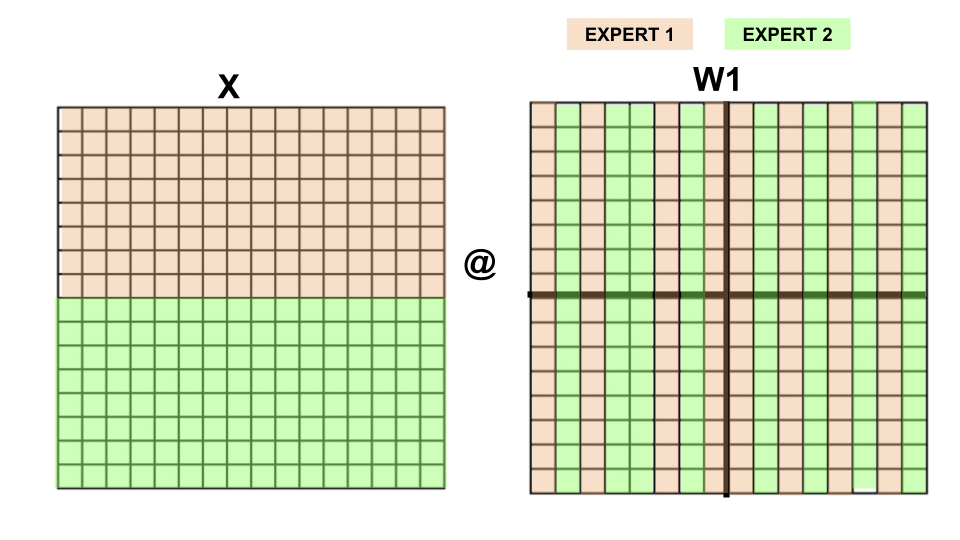}
        \caption{}
        \label{fig:moe3}
    \end{subfigure}
    \begin{subfigure}[b]{0.48\textwidth}
        \centering
        \includegraphics[width=\textwidth]{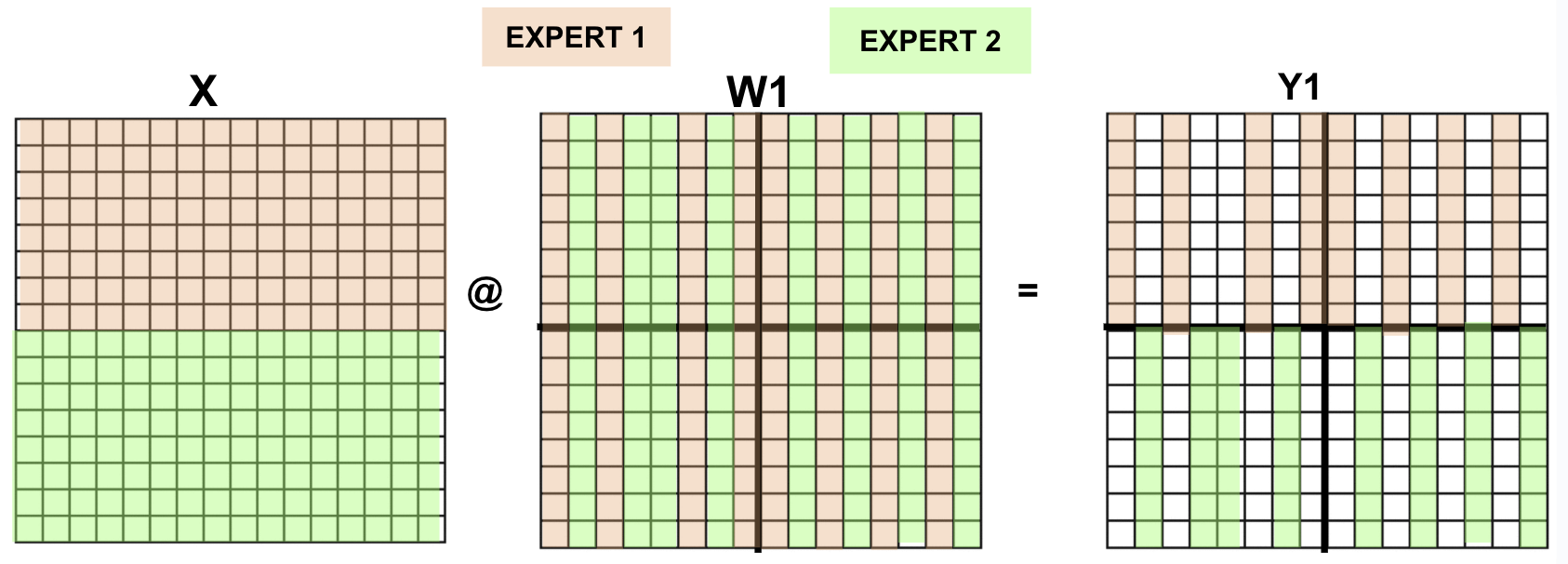}
        \caption{}
        \label{fig:moe4}
    \end{subfigure}
    \caption{\centering We describe our neuron level routing procedure step-by-step here. (a).The columns of $W_1$ are clustered. Each expert corresponds to a cluster center and columns are routed by L2 distance. (b) The tokens correspond to the rows of X. Each token is routed to an expert. The rows are colored accordingly. (c) Rows are permuted so that rows corresponding to tokens that are routed to the same expert are stacked together. (d) Only the rows and columns that have the same color are multiplied.  }
    \label{fig:moe_main}
\end{figure}

\subsection{Neuron Level Experts To Achieve Venom Format }\label{design:act}
Models trained with SquaredReLU activation function naturally exhibit $90\%$ sparsity, which we want to leverage with the Venom format. Naively pruning elements from the activations in a static pattern results in a severe loss degradation. This is because unlike the weights, the activations are input dependent, and the sparsity percentage of each feature varies depending on the sequence. Thus, we use a router that selects which specific features are kept for each token. We then reorder the tokens within the batch so that consecutive tokens have the same sparsity pattern and the following matrix multiplication be hardware accelerated: this is the Venom format. The exact algorithm for this is detailed in Algorithm~\ref{alg:Venom} and Figure~\ref{fig:moe_main}. First, in an offline phase (during the warmup dense steps), $W_1$'s columns are clustered into non overlapping subsets of columns. Then, the following procedure is applied during dense training steps: 
\begin{enumerate}
    \item Determine which experts to route each token to: Compute ($x @ ExpertMeans$ ) to get per token experts (Figure~\ref{fig:moe2}).
    \item Permute the tokens so that tokens routed to the same expert are placed in adjacent rows (Figure~\ref{fig:moe3}).
    \item Perform a batched matrix multiplication where a batch corresponds to the tokens routed to a given expert. 
    i.e
   $\begin{bmatrix} x_{s1}@W_{e1} \\x_{s2}@W_{e2} \\ x_{s3}@W_{e3}\\...\\ x_{sn}@W_{en}  \end{bmatrix}$. Here, $x_{sk}$ indicates subset of rows that are routed to expert $k$. The resulting matrix will be in the V:N:M format.
\end{enumerate}

\begin{center}
\begin{algorithm}[ht]
    \caption{Convert Activation to Venom}
    \label{alg:Venom}
    \begin{algorithmic}[1]
    \STATE x: [B, d]\\

    out = zeros([B, d])\\
    \FOR{$i = 1$ to $B$}
        \STATE expertids = columns with min cosine distance
        \FOR{expertid $= 1$ to expertids}
        \STATE {out[i] $+=$  x[i] @ W[expertid]}
        \ENDFOR
    \ENDFOR

    \end{algorithmic}
\end{algorithm}
\end{center}
\subsection{Theoretical End-to-End Speedup Analysis}\label{sec:roofline}
We show that our sparsification methods result in end-to-end speedups through theoretical FLOPs estimates and a roofline analysis. The total number of training FLOPs $= 6 \times \operatorname{numtokens} \times \operatorname{parameter count} = 6 BT(3DF + 2D(N+K)H)L$ where $B$ is the batch size, $T$ the sequence length, $D$ the embedding dimension, $L$ the network depth, $F$ the MLP hidden dimension, $N$ the number of query heads and $K$ the number of key value heads. The total FLOPs for the FFN block is $18BTDFL$. Hence, the ratio of FFN operations to total operations is $\frac{18BTDF}{6BT(3DF + 2D(N+K)H)}$. For large model sizes this can be up to $80\%$ of total floating point operations as illustrated in Figure~\ref{fig:flopfrac}. Because pretraining is generally compute bound, with communication overlapped with computation [\cite{hsia:isca2024}], this reflects the portion of the end-to-end time spent in the FFN. Thus any speedup in the FFN will result in a significant end-to-end speedup as well.

\subsubsection{Speedup from 2:4 sparse matrix multiplications}

The roofline for $2:4$ sparse matrix multiplications is $2$ times the roofline for regular matrix multiplication. However, our microbenchmarks in Section~\ref{sec:exp_microbenchmarks} show that in reality we can observe around a $1.5\times$ speedup. This amounts to an end-to-end speedup of $1.2\times$ if we apply $2:4$ sparsity to all the matrix multiplications in the FFN block for the 405B scale models (without GQA). GQA would increase this further.
In order to perform matrix multiplications on $2:4$ matrices, we need to first convert the matrix to the $2:4$ format: decide which elements to set to zero, and store the matrix in a packed format along with sparsity metadata. We see in Section~\ref{sec:exp_microbenchmarks} that this overhead is negligible. This involves a element-wise scan of the matrix which is memory bound, as it involves loading elements of matrix $A$ and storing the result back to global memory. When training with pipeline parallelism, this is a one time negligible cost, amortized over the number of microbatches processed.

\subsubsection{Speedup from Venom matrix multiplications}
Our microbenchmarks (Section~\ref{sec:exp_microbenchmarks}) show that Venom sparse GEMMs are up to $7\times$ faster than dense matrix multiplications. If we plug this into the expression for the ratio of FFN FLOPs to total FLOPs, we get around a $2.6\times$ end-to-end speedup for $1B$ and $7B$ scale models and a $4.2\times$ speedup for 405B scale models. We also need to factor in the overhead of converting data to the Venom format:
\begin{enumerate}
    \item Step 1: Token/expert routing. This is a matrix multiplication of the tokens with the experts mean. Given that the number of experts is small (typically 16), this operation is memory bound on reading the $[B, D]$ input, which should be negligible for larger models with a large model dimension.
    \item Step 2: Permutation of the rows. This is again memory bound on reading the input. While this step is necessary on Hopper GPUs, Blackwell introduces Scatter/Gather GEMM with TMA support at the hardware level. This eliminates this step entirely.
    \item Step 3: This involves a batched matrix multiplication, which will be compute bound for typical matrix shapes used during training.
\end{enumerate}
Fp8 quantization is an orthogonal optimization that can boost the end-to-end speedup of both 2:4 and Venom sparsity. This is because the preprocessing/sparsification of the matrices can be fused with quantization (especially during operations involving elementwise scans).

\section{Experimental Methodology}\label{sec:experiment}

\begin{table*}[ht]
\centering
\caption{\centering Model architecture configurations.}
\begin{tabular}{|l|l|l|l|l|l|l|l|l|}
\hline
                    & \textbf{num\_layers} & \textbf{d\_model} & \textbf{d\_ffn} & \textbf{num\_heads} & \textbf{batch\_size} & \textbf{seq\_len} & \textbf{lr} & \textbf{lr scheduler} \\ \hline
\textbf{Llama 1B} & 22                   & 2048              & 8192            & 16                  & 2                    & 8192              & 3e-4        & cosine                \\ \hline
\textbf{Llama 7B}   & 32                   & 4096              & 16384           & 32                  & 2                    & 8192              & 4e-4        & cosine                \\ \hline
\end{tabular}
\label{tab:modelarchs}
\end{table*}

 We run all our training experiments on Nvidia-H200 GPUs. We train Llama-3 model variants at 1B and 7B scale. The model architectures are presented in Table~\ref{tab:modelarchs}. We train on the DCLM dataset and perform evaluation on several benchmark tasks in the Eval-Harness suite [\cite{eval-harness}]. We compare against dense Llama-3 1B and 7B trained without any sparsity as a baseline. We compare and contrast the training loss curves (averaging the results of the last 100 steps as the final loss) and the evaluation accuracy. 
 
We empirically evaluate the efficacy of our sparsity methods on both accuracy and theoretical speedup. We start with the main result in Section ~\ref{sec:exp:main_result}. Our best recipe involves training using a combination of sparse and dense steps. The optimal combination of sparse:dense is 1:1 for the Llama 1B model and 1:3.5 for the Llama 7B model. 
 
 We subsequently present the ablation studies experiments in Section~\ref{sec:exp:ablation} to demonstrate how we arrived at this recipe and discusses general trends and conclusions we drew from this study. We study the impact of weight sparsification (using soft-thresholding) on each of the individual weights ($W_1$ and $W_2$) as well as the impact of weight sparsification on the transposition of the weight matrices. We then present activation sparsification ablation and look at the impact of applying 2:4 activation sparsity as well as Venom sparsity on the activations on accuracy. We see from the experiments that sparsification does result in slight accuracy/training loss degradation. One way to overcome this and recover the accuracy is to do a mixture of sparse and dense training steps. We investigate the order and and the ratio of sparse and dense steps to effectively recover accuracy in Section~\ref{sec:exp:dense_sparse}. Finally, in Section~\ref{sec:exp_microbenchmarks}, we present microbenchmarks for the 2:4 sparse matrix multiplications and Venom multiplications that we used to estimate end-to-end speedups.
 
\section{Evaluation Results}
\subsection{Result Overview}\label{sec:exp:main_result}

\begin{table*}[ht]
\centering
\caption{\centering Evaluation performance and training loss for Llama 1B and Llama 7B models using our best recipe. For the 1B model, the best configuration consists of 30k sparse and 30k dense steps. For the 7B model, the best configuration consists of 10k sparse and 38k dense steps.}

\begin{tabular}{|c|c|c|c|c|c|c|c|c|c|}
\hline
                                                             & \begin{tabular}[c]{@{}l@{}}\textbf{Speedup}\end{tabular} & \textbf{Loss}  & \textbf{Arc Easy} & \textbf{Arc Challenge} & \textbf{Piqa}  & \textbf{Obqa}  & \textbf{Hellaswag} & \textbf{Winogrande} & \textbf{Average} \\ \hline
\begin{tabular}[c]{@{}l@{}}Llama 1B\\ Baseline\end{tabular}  & 1.0                                                            & 2.758 & 0.602    & 0.323         & 0.741 & 0.362 & 0.563     & 0.571      & 0.527   \\ \hline
\begin{tabular}[c]{@{}l@{}}Llama 1B \\ Sparse\end{tabular}   & 1.352                                                          & 2.755 & 0.608    & 0.325         & 0.730 & 0.346 & 0.562     & 0.590      & 0.527   \\ \hline
\begin{tabular}[c]{@{}l@{}}Llama 7B \\ Baseline\end{tabular} & 1.0                                                            & 1.863 & 0.737    & 0.468         & 0.799 & 0.43  & 0.769     & 0.7        & 0.651  \\ \hline
\begin{tabular}[c]{@{}l@{}}Llama 7B \\ Sparse\end{tabular}   & 1.387                                                          & 1.866 & 0.738    & 0.464         & 0.793 & 0.46  & 0.764    & 0.691      & 0.652  \\ \hline
\end{tabular}
\label{tab:1}
\end{table*}

Table ~\ref{tab:1} presents the training loss and evaluation accuracy results for the recipe that we found to achieve the most speedup while recovering accuracy for $1B$ and $7B$ scale models respectively. Concretely, we train using the combination of of weights and activation sparsity detailed in Section ~\ref{sec:proposed_design}. This allows us to accelerate all 6 matrix multiplications in the FFN. We use Venom for activation sparsification and soft-thresholding for weight sparsification. On the 1B model, we train for 30k dense and 30k sparse steps. On the 7B scale model, we train for 10k sparse and 38k dense steps. From the calculations in Section ~\ref{sec:proposed_design}, we see that this amounts to a speedup of $2.2\times$ per iteration. Thus, with half the number of sparse iterations, the end to end speedup is $1.37\times$ for the 1B case. Table ~\ref{tab:1} present the average loss over the last $100$ steps and the evaluation results. We see that in both the cases our optimal recipe recovers training loss and evaluation scores while providing upto $1.37\times$ end-to-end speedup.  

\subsection{Ablation Studies}\label{sec:exp:ablation}
\subsubsection{Weight sparsification ablations}\label{sec:exp:weight_ablation}
Although our final recipe involves combining different weight and activation sparsities, understanding the effect of individual sparsification can help build intuition on how we arrived at the final combination recipe.

\begin{figure}[th]
    \centerline{\includegraphics[width=0.5\columnwidth]{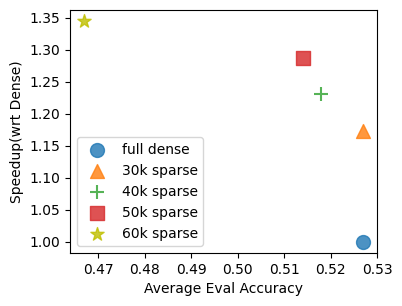}}
    \caption{\centering Pareto-frontier of model accuracy and speedup performance for the Llama 1B model.
    }
    \label{fig:paretofrontier1b}
\end{figure}

\begin{table}[ht]
\centering
\caption{Table presenting ablations for weight and activation sparsification. Various combinations of weights and activations are progressively sparsified. These are trained with only sparse steps.}
\begin{tabular}{|l|c|}

\hline
                        & \textbf{Training Loss} \\ \hline
Baseline (dense)        & 2.758                  \\ \hline
W1 sparse               & 2.784                  \\ \hline
W1, W1.T sparse         & 2.798                  \\ \hline
W2 sparse               & 2.785                  \\ \hline
W2, W2.T sparse         & 2.799                  \\ \hline
W1, W2 sparse           & 2.844                 \\ \hline
2:4 activation sparsity & 2.755                  \\ \hline
Venom sparsity          & 2.789                  \\ \hline
\end{tabular}
\label{tab:21}
\end{table}

Table~\ref{tab:21} presents the weight sparsification ablations. For these ablations, we do not train on a combination of sparse and dense steps. We train for all the steps using sparse GEMMs. Sparsifying each individual weight ($W_1$ or $W_2$) using soft-thresholding causes approximately 0.03 degradation in training loss. However, sparsifiying the transpose of the weights, in addition, to the weights results in  0.015 further loss degradation. However, there is not a significant difference in loss when sparsifying $W_1$ versus $W_2$. When sparsifying both $W_1$ and $W_2$, the total degradation in loss is additive.

Table~\ref{tab:21} also presents the results, illustrating the effect of activation sparsification on accuracy and training loss. We see that 2:4 sparsification with just $80\%$ results in nearly identical training loss and accuracy. Applying Venom sparsification results in approximately 0.03 degradation in loss (similar to the effect of weight sparsification). However, applying Venom sparsified matrix multiplications gives us $6\times$ speedup so this tradeoff is worth it, especially if we can recover the accuracy with additional dense finetuning steps.


\subsubsection{Recipe for training both Sparse and Dense Steps} 
\label{sec:exp:dense_sparse}

\begin{table*}[ht]
\centering
\caption{\centering Benchmark results from applying different combinations of sparse and dense steps.}
\begin{tabular}{|c|c|c|c|c|c|c|c|}
\hline
\textbf{\begin{tabular}[c]{@{}l@{}}Number of \\ Sparse Steps\end{tabular}} & \multicolumn{1}{l|}{\textbf{arc\_easy}} & \multicolumn{1}{l|}{\textbf{arc\_challenge}} & \multicolumn{1}{l|}{\textbf{piqa}} & \multicolumn{1}{l|}{\textbf{obqa}} & \multicolumn{1}{l|}{\textbf{hellaswag}} & \multicolumn{1}{l|}{\textbf{winogrande}} & \multicolumn{1}{l|}{\textbf{Average}} \\ \hline
\textbf{0 (Baseline)}                                                      & 0.602                                   & 0.323                                        & 0.741                              & 0.362                              & 0.563                                   & 0.57142                                  & 0.527                                 \\ \hline
\textbf{30000}                                                             & 0.608                                   & 0.325                                        & 0.730                              & 0.346                              & 0.562                                   & 0.590                                    & 0.527                                 \\ \hline
\textbf{40000}                                                             & 0.604                                   & 0.308                                        & 0.727                              & 0.346                              & 0.556                                   & 0.567                                    & 0.518                                 \\ \hline
\textbf{50000}                                                             & 0.6                                     & 0.308                                        & 0.721                              & 0.348                              & 0.544                                   & 0.565                                    & 0.514                                 \\ \hline
\textbf{60000}                                                             & 0.590                                   & 0.303                                        & \multicolumn{1}{l|}{0.714}         & 0.348                              & 0.532                                   & 0.561                                    & 0.467                                 \\ \hline
\end{tabular}
\label{tab:4}
\end{table*}

Table~\ref{tab:4} presents results varying the number of sparse and dense steps. Empirically, we found that sparse steps in the beginning followed by dense steps to recover accuracy works best. However, we begin Venom sparsification after a warmup of 1000 steps. This is because naturally activation sparsity takes form after that. We see that we need to train densely on half the number of steps in order to recover the accuracy. Figure~\ref{fig:paretofrontier1b} presents the pareto-frontier plot for the 1B model. We see that we are able to recover accuracy while getting speedups.

\subsubsection{Microbenchmarks}\label{sec:exp_microbenchmarks}

\begin{figure}[th]
\centerline{\includegraphics[width=0.5\columnwidth]{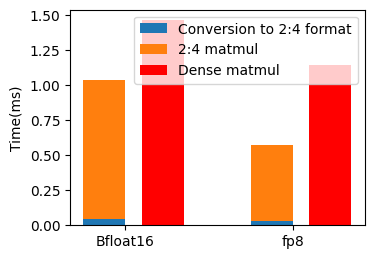} }
\caption{\centering Microbenchmarks comparing $2:4$ matrix multiplications and overheads of $2:4$ sparsification with a dense matrix multiplication baseline.}
\label{fig:24benchmarks1}
\end{figure}

\begin{figure*}[th] 
    \centering
    \begin{subfigure}{0.32\textwidth}
        \centering
        \includegraphics[width=\textwidth]{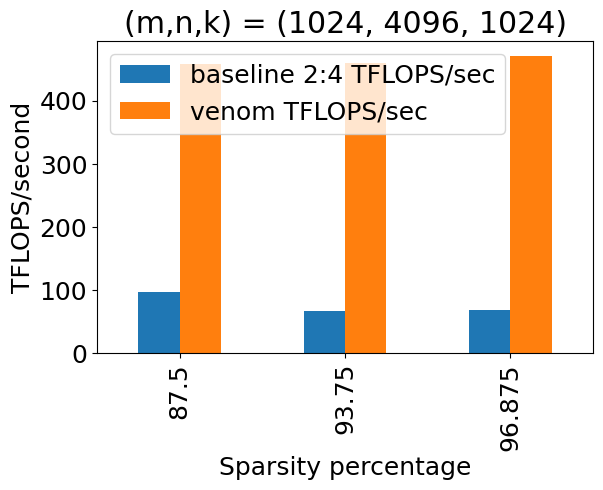}
        \label{fig:sub1}
    \end{subfigure}
    \begin{subfigure}{0.32\textwidth}
        \centering
        \includegraphics[width=\textwidth]{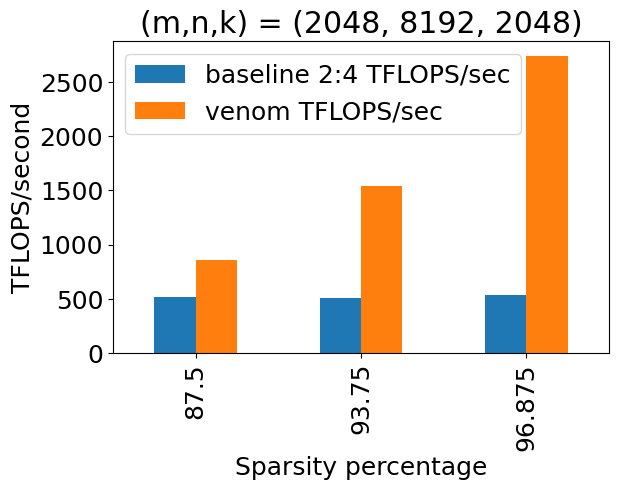}
        \label{fig:sub2}
    \end{subfigure}
    \begin{subfigure}{0.32\textwidth}
        \centering
        \includegraphics[width=\textwidth]{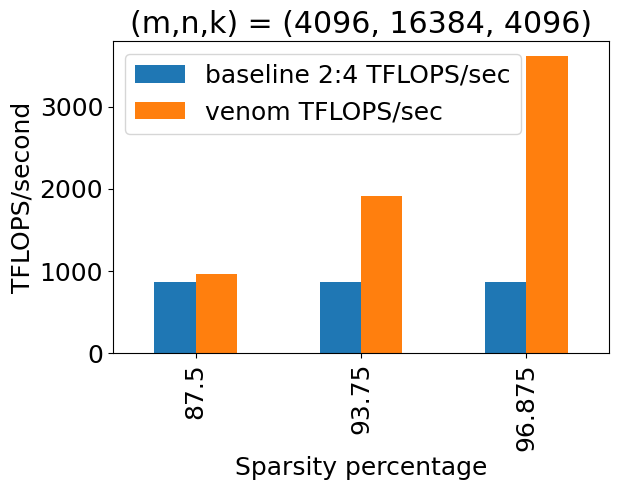}
        \label{fig:sub3}
    \end{subfigure}
    \begin{subfigure}{0.32\textwidth}
        \centering
        \includegraphics[width=\textwidth]{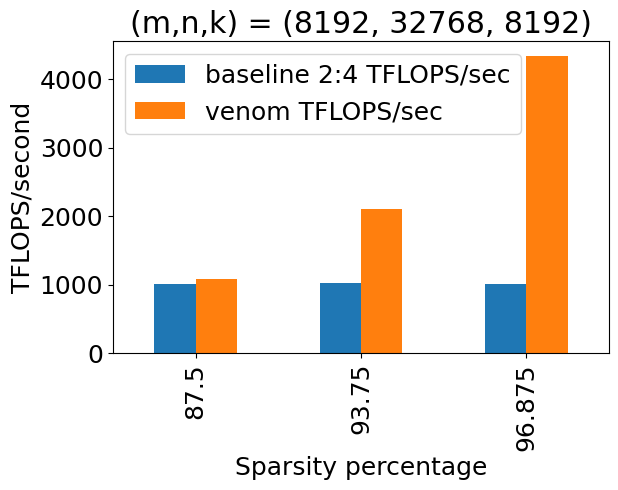}
        \label{fig:sub4}
    \end{subfigure}
    \begin{subfigure}{0.32\textwidth}
        \centering
        \includegraphics[width=\textwidth]{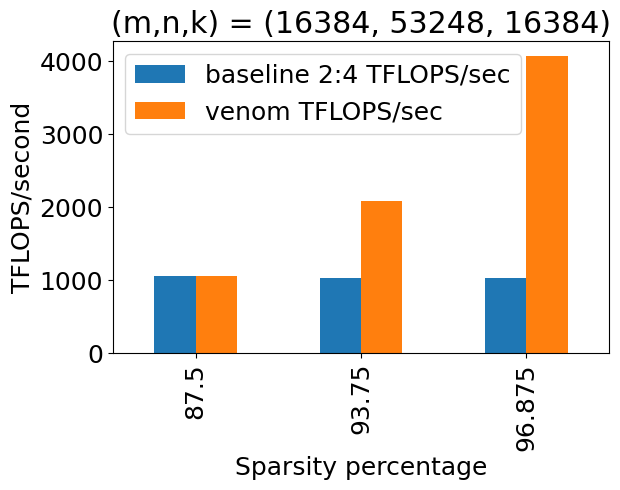}
        \label{fig:sub5}
    \end{subfigure}
    \caption{\centering Microbenchmarks that illustrate speedups when performing GEMMs on Venom matrices with different shapes.}
    \label{fig:Venom}
\end{figure*}

\begin{table}[ht]
\centering
\caption{\centering Different values of M give rise to different sparsities.} 
\begin{tabular}{|c|c|}
\hline
\textbf{V, N, M} & \textbf{Sparsity\%} \\ \hline
64, 2, 16        & 87.5                \\ \hline
64, 2, 32        & 93.75               \\ \hline
64, 2, 64        & 96.875              \\ \hline

\end{tabular}
\label{hyper:Venom}
\end{table}

Figure~\ref{fig:24benchmarks1} presents microbenchmarks on applying 2:4 soft-thresholding weight sparsity and 2:4 matmul for matrices of size $(m,n,k) = (4096, 16384, 4096)$. We see that speedup varies depending on the matrix size but we see upto $1.5\times$ speedup. We see that conversion to the 2:4 format using soft-thresholding presents a negligible overhead. Figure ~\ref{fig:Venom} presents microbenchmarks for Venom sparsification and multiplication. We see that the speedups here vary depending on the sparsity percent (and subsequently on hyperparameters $V$ and $M$). Table~\ref{hyper:Venom} displays the values of $V$, $M$ and $N$ to get the levels of sparsity observed. These parameters determine the sparsity levels. We see overall that a $6-8\times$ speedup can be expected.

\section{Related Work}
\textbf{Sparsity in LLMs.} 
Prior literature explored sparsity in weights and activations in deep neural networks even 
before the proliferation of modern Transformer style LLMs [\cite{hoefler2021sparsitydeeplearningpruning}]. In order to observe sparsity in weight matrices, weights must be pruned (set to zero). There are different techniques for weight pruning from magnitude based pruning methods to pruning methods that take higher order information (gradient and hessian) while deciding which weights to prune [\cite{sun2024simpleeffectivepruningapproach}]. Certain activation functions such as ReLU give rise to natural sparsity, i.e it sets elements less than or equal to zero to zero. This sparsity is different from weight sparsity in that it is naturally occurring and input dependent (dynamic).  \cite{li2023lazyneuronphenomenonemergence} and \cite{liu2023dejavucontextualsparsity} studied this in detail. Sparsification can be applied in different parts of the training pipeline from pre-training to downstream inference. It can be applied to the FFN or attention in Transformers. Sparse attention remains challenging particularly in the long context scenario since dropping elements results in loss of contextual information. However, some methods propose dropping the weights with the smallest magnitude.
Typically, sparsification was performed as a finetuning or post-training step with the aim of accelerating inference. 

\textbf{Structures of sparsity and hardware/systems support.} 
In structured sparse formats, zeros occur in squares/rectangles or a combination of the two (semi-structured). Various different formats can be categorized as semi-structured sparsity, including 2:4 sparsity [\cite{mishra2021accelerating}], v:n:m sparsity [\cite{Venom}] which combines 2:4 sparsity with column pruning. 
Effectively utilizing sparsity involves getting speedups commensurate with the number of zeros. Ideally, whenever there is a zero, one can save FLOPs by not performing the computation. Unfortunately, existing hardware doesn't have this kind of support for arbitrary sparsity patterns. Thus, there exists a variety of works that show how to efficiently and effectively transform unstructured sparse matrices to structured or semi-structured matrices by (1) permutations [\cite{pitpaper}] or by decomposing a matrix into a sum of (semi) structured sparse matrices [\cite{setty2025blockencodingsparsematrices}].

\textbf{2:4 sparsity.} 
Various works explored accelerating LLMs using them in all the stages (pre-training, fine-tuning and inference). For inference, \cite{fang2024maskllm} propose to learn a pruning mask that prunes the matrices so that they are in the 2:4 format. Some methods suggest finetuning using 2:4 sparsified weights so that one can accelerate inference (since the final model will be 2:4 sparse) [\cite{yang2024s2ftefficientscalablegeneralizable}]. Sparsification during training presents several systems and modeling challenges. \cite{jeong2025enablingunstructuredsparseacceleration} propose decomposing a matrix into the sum of several random structured sparse matrices (inspired by Taylor Series expansions). They also introduce novel hardware techniques to accelerate this. During training, sparsification must be done on-demand at every iteration. This could incur overheads if not carefully mitigated.  \cite{pytorch_accelerating_training_2023} explore systems technique to accelerate training BERT and vision models using 2:4 sparsity such as ensuring that the matrix and its transpose are sparse, kernel tiling using efficient memory I/O and using sorting networks to sort elements to decide which ones to prune. Weight sparsification presents quality challenges since the loss function can become discontinuous since the pruning mask can change at every iteration. Various methods have been proposed to mitigate this issue. \cite{hu2024acceleratingTransformerpretraining24} propose adding a regularizer term in the loss function. \cite{hu2024sstecontinuouspruningfunction} show that soft-thresholding is an effective technique that provably makes the loss function continuous and results in a stable loss curve. \cite{haziza2025acceleratingtransformerinferencetraining} propose using 2:4 sparsity on activations by using the squared ReLU activation function to accelerate the matrix multiplications involving activations in the FFN part of Transformers.

\section{Conclusion}
We explored a novel activation function that learns a neuron level router to give us a matrix that is in Venom format. We showed that using this format can accelerate the matrix multiplications pertaining to the activations by more than $2\times$. Combined with weight sparsity using soft-thresholding, this gives us a recipe to accelerate matrix multiplications in the FFN blocks in Transformers while recovering accuracy. In addition, we investigate and present the optimal combination of sparse and dense steps to recover accuracy. 

\section{Acknowledgment}
We would like to thank Bilge Acun, Chien-Yu Lin, Jeff Johnson, Matthijs Douze, Herve Jegou, and Mostafa Elhoushi for discussions that have helped shape this work.

\clearpage
\newpage
\bibliographystyle{assets/plainnat}
\bibliography{paper}



\end{document}